# Confluence: A Robust Non-IoU Alternative to Non-Maxima Suppression in Object Detection

Andrew J. Shepley, Greg Falzon, *Member, IEEE*, Paul Kwan, *Senior Member, IEEE* and Ljiljana Brankovic

**Abstract**—*Confluence* is a novel non-Intersection over Union (IoU) alternative to Non-Maxima Suppression (NMS) in bounding box post-processing in object detection. It overcomes the inherent limitations of IoU-based NMS variants to provide a more stable, consistent predictor of bounding box clustering by using a normalized Manhattan Distance inspired proximity metric to represent bounding box clustering. Unlike Greedy and Soft NMS, it does not rely solely on classification confidence scores to select optimal bounding boxes, instead selecting the box which is closest to every other box within a given cluster and removing highly confluent neighboring boxes. Confluence is experimentally validated on the MS COCO and CrowdHuman benchmarks, improving Average Precision by up to 2.3-3.8% and Average Recall by up to 5.3-7.2% when compared against de-facto standard and state of the art NMS variants. Quantitative results are supported by extensive qualitative analysis and threshold sensitivity analysis experiments support the conclusion that Confluence is more robust than NMS variants. Confluence represents a paradigm shift in bounding box processing, with potential to replace IoU in bounding box regression processes.

---------- ◆ ----------

## 1 INTRODUCTION

OBJECT detection algorithms that rely on Deep Convolutional Neural Networks (DCNNs) learn data-derived features that allow objects of interest to be localized in images [1]. State-of-the-art DCNNs return dense clusters of bounding boxes of varying sizes and locations, which congregate in regions likely to contain an object [2],[3],[4],[5], as illustrated in Fig. 1 and Fig. 2. To obtain the final detection output, only one optimal bounding box per object must be retained, and all excess bounding boxes must be removed. In cases where objects occlude each other, overlapping bounding boxes should be retained so that each object is represented by one bounding box.

This task is usually performed by variants of the Non-Maxima Suppression (NMS) algorithm [6],[7]. NMS uses the classification confidence score attributed to each bounding box by the DCNN to sort the bounding boxes in descending order. The highest scoring box is selected as the 'optimal' box for the first object. NMS then employs a user-defined Intersection over Union (IoU) threshold to suppress or decay the confidence score of all bounding boxes whose overlap with the 'optimal' box exceeds the threshold. This results in the removal of boxes that overlap heavily with the 'optimal' box. If there is more than one object in the image, NMS then selects the next highest scoring bounding box in the remaining set as the best box to represent the next object and repeats the suppression procedure. This process is repeated until the final detection set is attained – optimally, one bounding box per object.

NMS usually achieves acceptable performance in images where objects do not overlap each other much

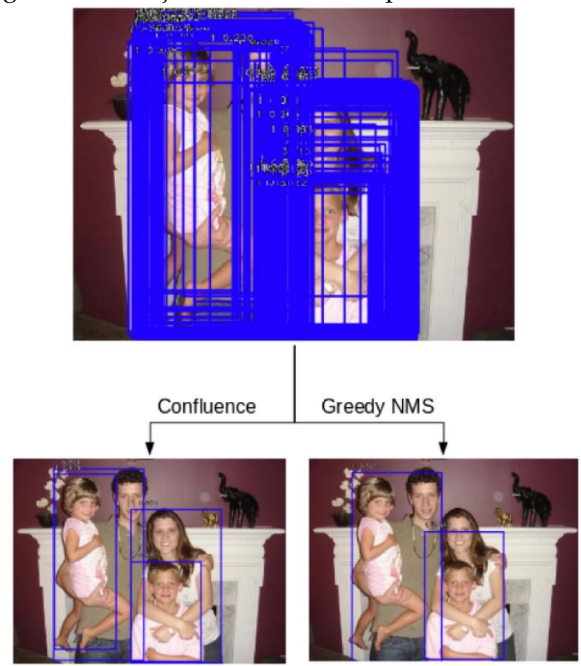

Fig. 1. Confluence accurately selects an optimal bounding box for each person, with no retention of false positives. NMS selects a poorly localized maxima (confidence score: 96.4%), resulting in two optimal boxes surrounding the woman and boy (85.9% and 88.2% respectively) being suppressed by the IoU threshold. Due to high IoU, the bounding box surrounding the girl is also suppressed by NMS, reducing recall.

---

- *A.J. Shepley is a postdoctoral researcher at the School of Science and Technology, University of New England, Armidale, NSW, Australia. E-mail: asheple2@une.edu.au.*
- *G. Falzon is an Associate Professor with Flinders University, Adelaide, SA, Australia. E-mail: greg.falzon@flinders.edu.au.*
- *Ljiljana Brankovic is a Professor at the School of Science and Technology, University of New England, Armidale, NSW, Australia. E-mail: ljiljana.brankovic@une.edu.au*
- *P. Kwan is a Professor at the Melbourne Institute of Technology, Sydney, NSW, Australia. E-mail: pkwan@mit.edu.au.*

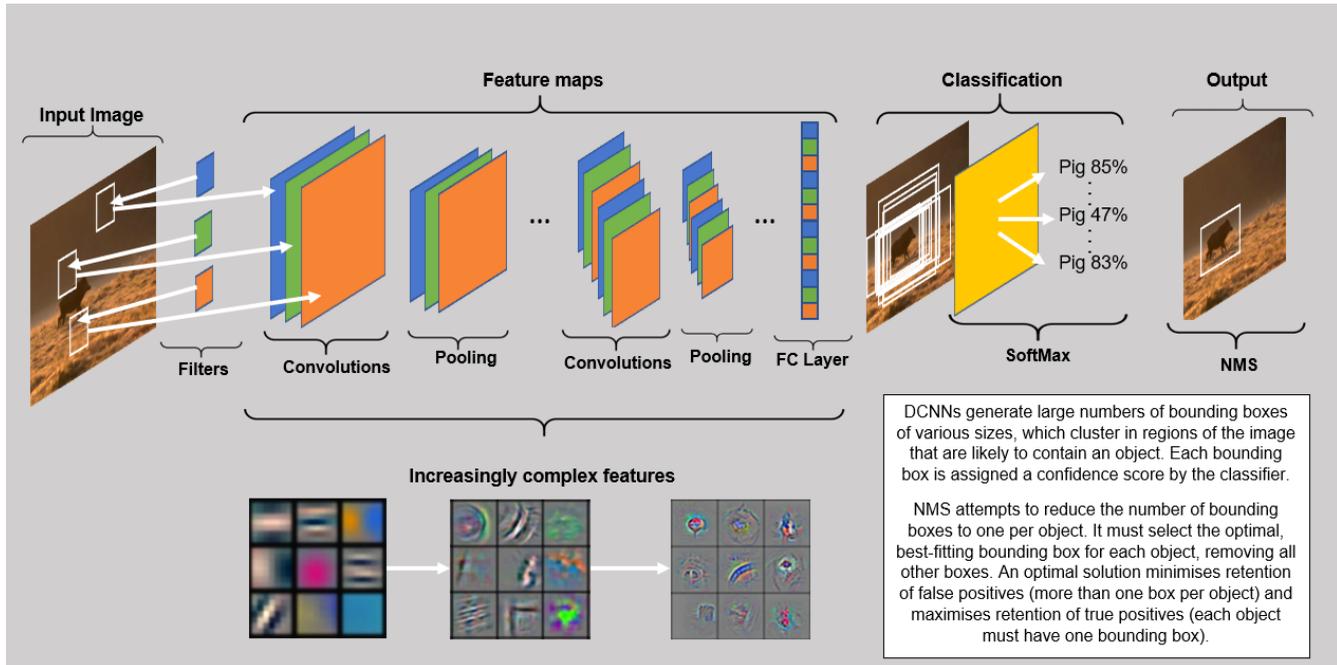

Fig. 2. Schematic diagram representing generic DCNN functionality fundamental to state-of-the-art object detectors such as Mask-RCNN [3]. FC refers to Fully Connected layers. Large numbers of bounding boxes are generated for each object, from which only one optimal bounding box should be retained. Diagram adapted from [82]. Images of complex features obtained from [83].

[8],[9]. However, when objects are occluded or overlap heavily such as in high density or crowded settings, reliance on IoU forces a trade-off between recall and precision [8],[9],[10],[11],[12]. This is because NMS is a heuristic algorithm that simply assumes that a high overlap between bounding boxes results in a high probability of one of the boxes being a duplicate [13]. Thus, to retain highly overlapping true positives, a higher IoU threshold must be used, leading to greater retention of false positives [14],[15],[16],[17],[18].

Another widely noted shortcoming of NMS is its poor localization accuracy occasioned by its sole reliance on the classification confidence score to select optimal bounding boxes [19],[20],[21],[22],[23],[24]. This score is a class label probability and does not effectively represent localization accuracy [24], exhibiting low correlation between localization accuracy and classification score [22]. Its usage by NMS means that NMS is susceptible to returning suboptimal bounding boxes, whilst suppressing better candidate boxes [21] as illustrated by Fig 1. It should therefore not be used as the primary metric by which the optimal boxes are selected.

Although alternatives to and variants of NMS have been proposed [7],[25],[26],[27],[21],[28],[29],[30],[17],[31], [12],[18],[15],[32],[33], the majority share the inner mechanisms of NMS [34], namely reliance on IoU and confidence scores. The most commonly used variants of NMS are Greedy NMS (G-NMS) [6], which is widely regarded as the de-facto standard solution [35],[36],[10], [37],[9],[14],[38],[39],[40],[22],[35],[41],[42],[43],[44],[30] and the recently proposed Soft NMS (S-NMS) [7]. S-NMS aims to reduce the greedy suppression of true positives by G-NMS by decaying rather than eliminating the confidence scores of highly overlapping boxes as a continuous function of their overlap with the optimal box. S-NMS is being increasingly adopted to replace G-NMS in object detection pipelines [45],[46],[47],[48],[49],[50], [51], [52],[53],[54] however it is also limited by its ongoing reliance on IoU and the classification confidence score.

Thus, this study presents Confluence, a novel non-IoU alternative to NMS algorithms. The key contributions are threefold. Firstly, we propose a proximity metric as an alternative to IoU in the suppression of false positives. Rather than using the IoU between bounding boxes to determine whether they represent the same object, we propose the confidence-weighted normalized pairwise Manhattan Distance [55] between corresponding bounding box coordinates to measure their coherence, and hence more accurately determine whether they point to the same object. Secondly, we propose Confluence NMS (NMS-C), a non-IoU variant of NMS, which retains bounding boxes using the classification confidence score, but suppresses false positives using the Confluence metric. Thirdly, we present Confluence, an algorithm which uses the proximity metric in both the selection of the optimal boxes, and the suppression of false positives. The effectiveness of Confluence and NMS-C is empirically validated against both G-NMS and S-NMS on the MS COCO [56] and CrowdHuman [57] datasets achieving gains in Average Precision of 2.3-3.8% and gains in Average Recall of 5.3-7.2%.

## 2 RELATED WORKS

Although NMS has been an algorithm of significant importance in computer vision for over 50 years, its shortcomings are as widely recognized as its essential role in the object detection pipeline [10],[58],[39],[40],[35],[11]. This has given rise to many adaptations of NMS, and various alternatives.

Many alternatives and adaptations aim to limit or eliminate the bottleneck caused by NMS in the object detection pipeline [59] by reducing computational expense and improving efficiency [60],[28],[61],[62],[63]. However, improvements in accuracy are achieved at the expense if performance, as these methods do not address localisation accuracy or recall [64].

Alternatively, some methods aim to improve proposal refinement during training to provide NMS with better input, thus improving performance [11],[24]. Other methods aim to circumvent NMS or eliminate the need for it altogether [65],[66],[67],[68],[69],[70],[9] however, they involve significant changes to neural network architecture, and achieve inferior or competitive performance [66]. This means they cannot be adopted in state-of-the-art object detectors [3],[5] that rely on NMS.

Some methods are specific to video [31],[71], body-part [72] or 3D [25],[73],[74] applications and are not applicable to or evaluated on standard object detection tasks. Other methods achieve gains in performance when compared to G-NMS and S-NMS but rely on additional data such as object depth information [17], or a combination of pixel-based and amodal bounding boxes [12], and corresponding network architecture changes, which are often not available or able to be implemented in standard object detection pipelines.

Most variants of NMS involve minor conceptual differences, resulting in similar performance. Matrix NMS was proposed by [64], which implements S-NMS in parallel for instance segmentation. Significant gains in speed were achieved, and it did outperform G-NMS but it did not outperform S-NMS and still relies on IoU and classification confidence scores. Similarly, [10] proposes GNet, a neural network that uses message passing between neighboring bounding boxes, whereby changes in bounding box representations are learned based on the 'negotiations' between bounding boxes to decide which bounding box will represent which object. Although this approach only uses bounding boxes and scores as input, it is a highly complex network which requires significant amounts of training data. In contrast, our approach does not require any training, and can be easily incorporated into systems that currently use NMS, without the need for architectural changes.

IoU-NMS [22] aims to improve bounding box localization by embracing IoU as the "natural criterion for localization accuracy". Initially, [22] proposes IoU-Net to learn the IoU relationship between ground truth box and the candidate boxes, returning a localisation confidence score per box. IoU-NMS then uses the localization confidence instead of the classification confidence score in the NMS procedure. Although [22] achieves higher Average Precision (AP) at higher IoU thresholds on several networks, it was not evaluated on standard benchmarks. Further, combining S-NMS with box-voting [75] returns similar results to [22] with reduced computational expense [21]. Notably, integration of IoU-Net in standard object detection pipelines is not easily achieved and requires retraining of object detectors, limiting its adoption.

Alternatively, a bounding box regression loss and a modified S-NMS dubbed Softer-NMS is proposed by [21], to improve bounding box localization. It avoids reliance on the classification confidence score by assigning a localisation score to each box, which like [22], is used to select the best box. The localisation score is however not sufficient to select the best box, with [21] relying on application of G-NMS or S-NMS followed by mean averaging of localization confidence scores of clustered boxes, and calculation of standard deviation to measure the uncertainty of the estimated bounding box location. Although [21] improved AP on MS COCO, its usage requires retraining of the object detector integrating the new loss function. It is also significantly more computationally expensive, and it only addresses the issue of poor localisation, and does not improve recall. Conversely, Confluence resolves both issues, by treating the strong coherence of bounding boxes as a measure of localization confidence without complex, computationally expensive calculations, retraining of networks or need of a localization loss.

Another limitation of NMS is its high sensitivity to the IoU threshold [37],[36],[35], which makes it difficult to achieve an optimal balance between retention of true positives and removal of false positives. This is particularly true for highly occluded settings such as crowds and has resulted in the development of crowd setting specific modifications of NMS. Rather than avoiding the use of IoU, these modifications are premised on attempting to minimise the limitations of IoU in adequately representing true and false positives.

Adaptive NMS [36] proposes a dynamic per instance IoU threshold to achieve a better balance between retention of true positives and elimination of false positives, tailored to pedestrian detection. A learnable sub-network determines the IoU threshold, which changes depending on the density of the image. It achieved competitive results on the CityPersons and CrowdHuman benchmarks. Alternatively, handcrafted image descriptors of person silhouettes are exploited by [35] to rescore candidate detections to improve performance of NMS on the PETS [76], COCO Person [56] and Okutama-Action [77] datasets. However, [36] and [35] are significantly more computationally expensive than traditional NMS and not evaluated in general object detection tasks.

Similarly [78] proposes attribute-aware NMS, which leverages semantic attributes such as density and diversity gained from a pedestrian-oriented attribute map to minimise false positives occasioned from using a high IoU threshold during NMS. Although it outperformed G-NMS, its reliance on a custom network designed specifically for crowd detection means it cannot be easily adopted in standard state-of-the-art object detection pipelines.

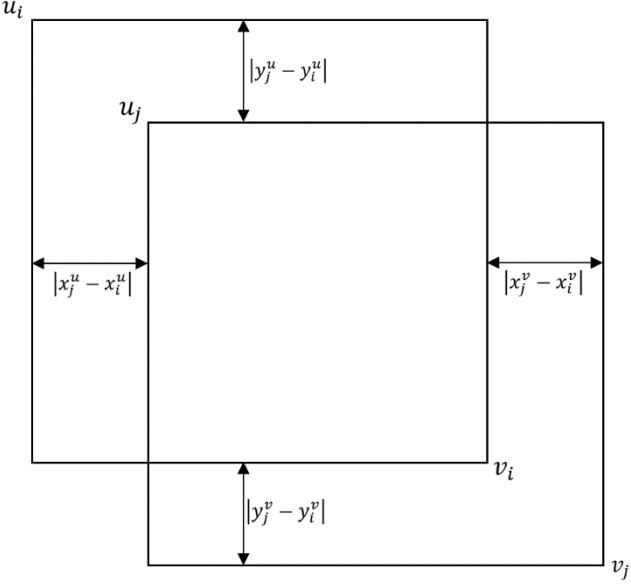

Fig. 3. Diagrammatic representation of the proximity calculation for two bounding boxes $b_i$ and $b_j$.

An adaptive threshold for NMS in video sequences is proposed by [71]. It uses high confidence bounding boxes in key frames to improve detection in lower scoring frames. Although gains in AP were achieved, it is not useful for standard image-based object detection due to its reliance on multiple consecutive frames for threshold adaptation. In contrast, Confluence can be used in both image object detection studies and video-based object detection without retraining or additional image information, achieving significant gains in performance.

## 3 METHODOLOGY

Confluence derives its name from the highly confluent, aggregated clusters of bounding boxes returned by a neural network when an object is detected. Rather than treating the excessive proposals as problematic, Confluence embraces them as a way of identifying and retaining the bounding box which best represents the object location. The clustering can be interpreted as a collective vote on object location, where the box that best represents every other
box is the optimal box. Bounding box confluence is also an effective way of removing those false positives that are confluent with the retained box.

Confluence is a recursive, two-staged algorithm which first retains an optimal bounding box, and then removes false positives that are confluent with it. Retention is achieved using a confidence weighted Manhattan Distance inspired proximity measure to evaluate bounding box coherence, enabling retention of the bounding box that best represents all boxes in a cluster. The second stage involves removal of all bounding boxes which are confluent with the retained bounding box. This process is repeated until all boxes have been processed.

### 3.1 Manhattan Distance

The Manhattan Distance (*MD*) or $L_1$ norm, is the sum of the vertical and horizontal distances between two points [55]. The *MD* between the points $u_i = (x_i^u, y_i^u)$ and $u_j = (x_j^u, y_j^u)$ is shown by equation (1):

$$MD(u_i, u_j) = |x_j^u - x_i^u| + |y_j^u - y_i^u| \qquad (1)$$

Each bounding box $b_i$ can be represented by two diagonally opposite corners. For example, $b_i = (u_i, v_i)$ can be defined by the upper left corner $u_i = (x_i^u, y_i^u)$ and the lower right corner $v_i = (x_i^v, y_i^v)$.

We propose a proximity measure $P(b_i, b_j)$ between any two bounding boxes $b_i = (u_i, v_i)$ and $b_j = (u_j, v_j)$, represented by the sum of the *MD* between the upper left corners $u_1 = (x_1^u, y_1^u)$ and $u_2 = (x_2^u, y_2^u)$, and the lower right corners $v_1 = (x_1^v, y_1^v)$ and $v_2 = (x_2^v, y_2^v)$ of the two boxes as given by equation (2):

$$P(b_i, b_j) = MD(u_1, u_2) + MD(v_1, v_2) \qquad (2)$$
$$= |x_2^u - x_1^u| + |y_2^u - y_1^u| + |x_2^v - x_1^v| + |y_2^v - y_1^v|$$

A small $P(b_i, b_j)$ value denotes highly confluent boxes $b_i$ and $b_j$, whilst a high $P(b_i, b_j)$ value indicates boxes that are not attributable to the same object - they may be somewhat overlapping, or completely disjoint. A diagrammatic representation of $P(b_i, b_j)$ is given in Fig 3.

Let $O(b_i)$ be a set of all boxes bounding the same object as box $b_i$, such that $O(b_i)$ does not include $b_i$ itself. We define proximity $P(b_i)$ of a box $b_i$ as the mean value of the proximities of the box $b_i$ to all the boxes in $O(b_i)$:

$$P(b_i) = \frac{1}{|O(b_i)|} \sum_{b_j \in O(b_i)} P(b_i, b_j) \qquad (3)$$

A bounding box $b_i$ surrounded by a dense cluster of bounding boxes will be characterized by very low $P(b_i)$ values, in comparison to a bounding box which is loosely surrounded by bounding boxes. The latter could be correctly categorized as an outlier, or as suboptimal. In effect, this provides a measure of the object detector's confidence in the presence of an object at a given location. On this basis, we propose that that the bounding box $b_i$ with the lowest $P(b_i)$, value represents the most confident detection for a given object.

Notably, this concept overcomes an issue faced by all variants that rely on the classification confidence score. In situations where the highest scoring bounding box is sub-optimal in comparison with another lower scoring bounding box, NMS returns the sub-optimal bounding box, as illustrated by Fig 1. In contrast, the $P(b_i)$ measure allows for the bounding box $b_i$ that is most confluent with all other bounding boxes assigned to a given object to be favored.

## 3.2 Normalization

The concept outlined in the previous section operates effectively in circumstances where bounding boxes are of similar size. However, in practice, objects and their corresponding bounding boxes will be of varying sizes. This poses a problem when regulating bounding box retention or removal using a hyper-parameter based on $P(b_i)$. This is because a trade-off between removing large false positives and retaining small true positives would need to be reached. For example, Fig. 4 shows two large bounding boxes on the right, which denote the same object. The two small bounding boxes on the left denote two separate objects. However, when the proximity calculation is performed on each pair of boxes, the same value is obtained, as follows:

$$P(b_1, b_2) = |3-2| + |3-4| + |4-3| + |5-6| = 4 \quad (4)$$

$$P(b_3, b_4) = |10-9| + |2-3| + |20-19| + |10-11| = 4 \quad (5)$$

This poses the problem of distinguishing between bounding boxes belonging to the same or distinct objects, particularly when there are significant differences in size.

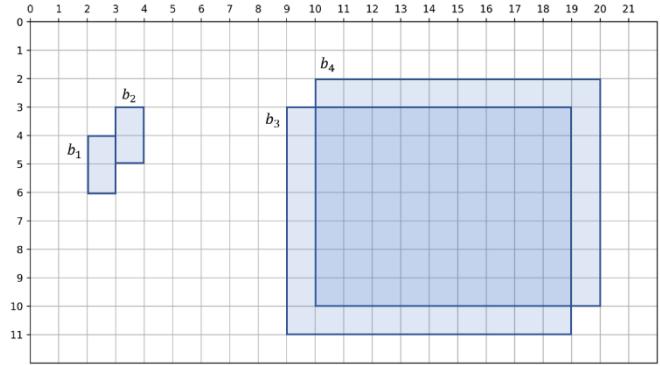

Fig. 4. Bounding box coordinates must be normalized, to ensure variations in bounding box size does not affect the P value, causing small disjoint boxes to have similar values to that of large highly overlapping boxes.

To overcome this issue, a normalization algorithm was used to scale the bounding box coordinates between 0 and 1, whilst preserving their relationship with each other. The normalization algorithm transforms each coordinate of the boxes $b_i$, and $b_j$ as follows.

$$b_i = (u_i, v_i) = ((x_i^u, y_i^u), (x_i^v, y_i^u))$$

$$b_j = (u_j, v_j) = ((x_j^u, y_j^u), (x_j^v, y_j^u))$$

$$X = \{x_i^u, x_j^u, x_i^v, x_j^v\}$$

$$Y = \{y_i^u, y_j^u, y_i^v, y_j^v\}$$

$$norm(x_i^u, y_i^u) = \left(\frac{x_i^u - \min(X)}{\max(X) - \min(X)}, \frac{y_i^u - \min(Y)}{\max(Y) - \min(Y)}\right) \quad (6)$$

The other relevant corners of the boxes $b_i$, and $b_j$ are normalized in the same way. Then we have:

$$norm(b_i, b_j)$$
$$= \left((norm(x_i^u, y_i^u), norm(x_i^v, y_i^v)), (norm(x_j^u, y_j^u), norm(x_j^v, y_j^v))\right) \quad (7)$$

Normalization allows intra-object and inter-object bounding boxes to be distinguished by making the relationship between large and small bounding boxes directly comparable. For example, after applying equation (6) to $b_1, b_2, b_3, b_4$ illustrated in Fig. 4, the co-ordinates belonging to $b_1$ and $b_2$ are transformed from $u_1 = (2,4), v_1 = (3,6), u_2 = (3,3)$ and $v_2 = (4,5)$ to $u_1 = (0,0.\dot{3}), v_1 = (0.5,1), u_2 = (0.5,0)$ and $v_2 = (1,0.\dot{6})$. When equation (2) is applied, a value of $1.\dot{6}$ is obtained, indicating the bounding boxes are not confluent. Similarly, the co-ordinates belonging to $b_3$ and $b_4$ are transformed from $u_1 = (9,3), v_1 = (19,11), u_2 = (10,2)$ and $v_2 = (20,10)$ to $u_1 = (0,0.\dot{1}), v_1 = (0.\dot{9}0,1), u_2 = (0.\dot{0}9,0)$, and $v_2 = (1,0.\dot{8})$. Reapplying equation (2) to these normalized co-ordinates returns a value of $0.\dot{4}0$, indicating high confluence. Thus, normalization allows the difference between intra-object and inter-object bounding boxes to be distinguished.

## 3.3 Intra-cluster Retention and Removal

As all bounding box pairs are normalized between 0 and 1, any pair of intersecting bounding boxes $(b_i, b_j)$ will have a $P(b_i, b_j)$ value below 2. However, empirical observation suggests that most clusters will be characterized by proximity values between 0 and 1.

In Fig. 5 and Fig. 6, each point on the horizontal axis represents a bounding box $b_i, 1 \leq i \leq n$, while the vertical axis represents the proximity $P(b_0, b_i)$ between a randomly selected box $b_0$, and box $b_i$. To visualize the relationship between boxes, they were ordered so that $P(b_0, b_1) \leq P(b_0, b_2) \leq \cdots \leq P(b_0, b_n)$. This reveals the blob-like nature of $P(b_i, b_j)$ clustering, where each horizontal blob represents an object. $P(b_i, b_j)$ values generally lie between 0 and 1, with the optimal bounding box being represented within the flattest gradient of the blobs.

Thus, if the $P(b_i, b_j)$ value of any two bounding boxes is below the user defined Confluence threshold ($C_t$), it is assumed that they belong to the same cluster, and therefore refer to the same object, or to one or more high density objects. The optimal intra-cluster bounding box is found, by calculating the mean $P(b_i)$ value, using equation (3), of each box in the cluster. The bounding box with the lowest $P(b_i)$ is the most confluent and is retained.

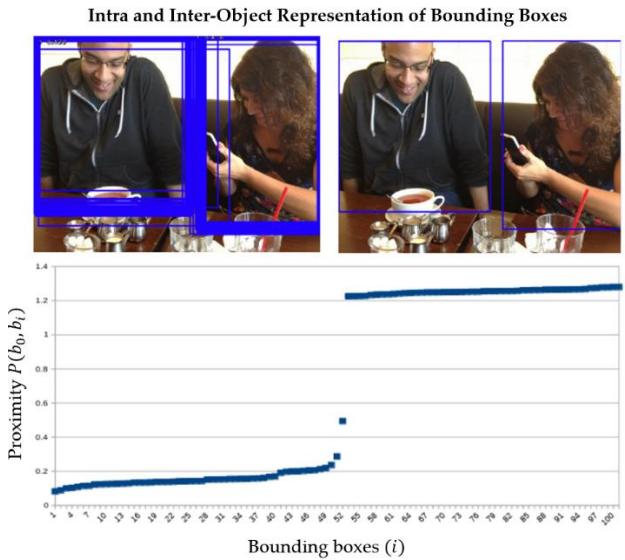

Fig. 5. Bottom left: Raw RetinaNet output, comprised of approx. 100 bounding boxes. Bottom right: Confluence output. Top: A bounding box $b_0$ was randomly selected from the raw RetinaNet output. The normalized proximity values between the selected box and every other box proposed by RetinaNet was calculated. These values were sorted in ascending order (to group proximate boxes) and plotted. This graph illustrates how the densest areas of bounding box confluence correspond to the sections of the scatter plot with flattest gradients. This pattern occurs regardless of which bounding box was randomly selected to start with.

All bounding boxes that are confluent with this chosen box are likely to be false positives. Thus, their classification confidence scores are either removed or decayed as a function of their confluence with the chosen box.

### 3.4 Confidence Score Weighting

The majority of NMS variants, including G-NMS and S-NMS use a single classification confidence score returned by the object detector as the sole means by which an 'optimal' bounding box is selected. In contrast, Confluence assesses the optimality of a given bounding box $b_i$ by comparing both its confidence score and its $P(b_i)$ values with competing bounding boxes. To achieve this, the $P(b_i)$ value is weighted by the classification confidence score $s_i$, as follows:

$$P_w(b_i) = P(b_i)(1 - s_i) \qquad (8)$$

As $s_i$ is a value which lies between 0.01 and 1 (all classification confidence scores lie between 1-100%), this in effect provides a bias in favour of high confidence boxes by artificially reducing the value of $P(b_i)$ (Note that all bounding boxes with confidence scores below 0.01 are not considered). Conversely, the $P_w(b_i)$ value of low confidence boxes will be greater. This increases the likelihood of a high confidence box being selected, as bounding boxes are chosen based on small $P_w(b_i)$ values.

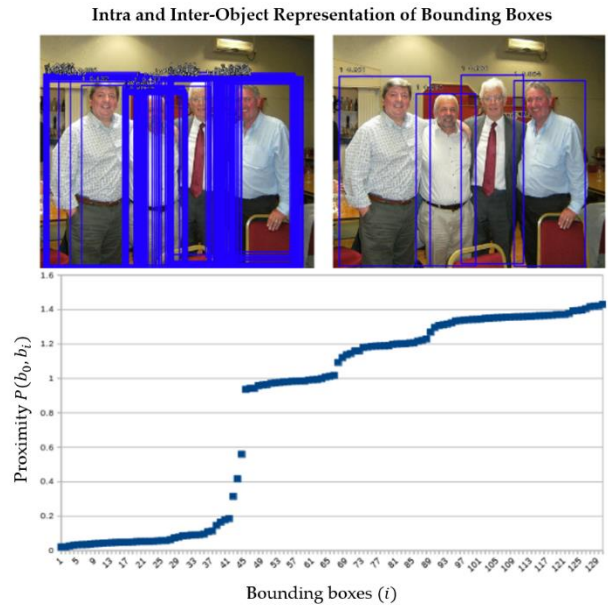

Fig. 6. Bottom left: Raw RetinaNet output, comprised of approx. 130 boxes. Bottom right: Confluence output. Even when objects are close together or overlapping, Confluence is capable of clearly distinguishing between intra-object and inter-object bounding boxes. For detailed information on how the graph was generated, see Fig. 5 caption.

This algorithm is based on the principle that a powerful classifier can be constructed by using the sum of weaker individual classifiers [79],[80]. Each individual $P_w(b_i)$ value is a weak classifier on its own, but when these weak classifiers are collectively interpreted, they provide a powerful means to classify a bounding box as either confident - via high confluence, or not confident, via disparate positioning with respect to other bounding boxes. In essence, this provides a vote of confidence by the object detector on which bounding box best represents every other bounding box assigned to an object. Our experimental results presented in Tables 1-3 suggest that this is a reliable means to accurately identify true positives, whilst effectively minimizing false positives. This allows achievement of optimal precision and recall values.

### 3.5 Implementation

The pseudocode outlining the Confluence is provided by Algorithm 1. Both Confluence and C-NMS were implemented in Python and are freely available on GitHub.[1]

---

[1] github.com/ashep29/confluence

**Algorithm 1** Confluence
**Input:** $I(b,s) = \{b_0 : s_0, .., b_n : s_n\}, C_t$
 $I$ is a dictionary mapping a box $b_i$ to its corresponding confidence score $s_i$.
 $C_t$ is the Confluence threshold
**Output:** $D$ is the final detection set
**begin:**
1: $C(b, P_w) \leftarrow \{\}$
 $C$ is a dictionary mapping each bounding box $b_i$ in $I$ to its corresponding weighted proximity $P_w(b_i)$
2: $N(b, neighbours) \leftarrow \{\}$
 $N$ is a dictionary mapping each bounding box $b_i$ in $I$ to a set of bounding boxes in $I$ that are confluent with $b_i$
3: $D \leftarrow \{\}$
4: **for** $b_i$ in $I$ **do**
5: $\sum P \leftarrow 0$
6: **for** $b_j$ in $I$ except $b_j = b_i$ **do**
7: $normalise(b_i, b_j)$
8: $P \leftarrow P(b_i, b_j)$
9: **if** $P < C_t$ **then**
10: $\sum P+ = P$
11: $N[b_i] \leftarrow N[b_i] \cup b_j$
12: **end if**
13: **end for**
14: $P_w = (\sum P/size(N[b_i])) * (1 - s_i)$
15: $C \leftarrow C \cup (b_i : P_w)$
16: **end for**
17: **while** $I \neq empty$ **do**
18: $b_m \leftarrow argmin(C)$
19: $D \leftarrow D \cup b_m, I \leftarrow I - b_m$
20: $I \leftarrow I - N[b_m]$
21: **end while**
22: **return** $D$

The main steps are as follows (bounding boxes are randomly ordered as order does not affect the final output):
1. For each bounding box in the input set $I$ (line 4):
   a. Create a variable $\sum P$ (line 5). This will be used to accumulate the normalized $P(b_i, b_j)$ values.
   b. Using the inner loop (lines 6-13), compare $b_i$ against every other box $b_j$.
   c. Use equation 7 to normalize the relationship between $b_i$ and every other bounding box $b_j$ (line 7).
   d. Calculate the normalized proximity between $b_i$ and every other bounding box $b_j$ (line 8).
   e. If the $P(b_i, b_j)$ value is below the user-defined $C_t$ threshold, increment $\sum P$ (line 10) by $P(b_i, b_j)$ and add $b_j$ to the set of neighbors of $b_i$, i.e., $N[b_i]$ (line 11).
   f. Once $b_i$ has been compared to every other bounding box, calculate its $P_w(b_i)$, and add the item $(b_i: P_w(b_i))$ to the dictionary $C$ (line 15).
2. Once all boxes in $I$ have been processed, identify the bounding box with the lowest $P_w$ value ($b_m$) (line 18).
3. $b_m$ is the optimal bounding box, which is added to the final detections set $D$ and removed from the dictionary $I$ (line 19).
4. Next, remove all bounding boxes that have been identified as neighboring $b_m$ as these are treated as locating the same object.
5. Repeat steps 2-4 until all boxes have been processed.

The computational complexity of each step of Confluence is $O(n)$, where the size of the input set of bounding boxes is $n$. This is due to the calculation of the normalized proximity score. As this measure is computed for each bounding box against every other bounding box, Confluence has an overall computational expense of $O(n^2)$.

Although the computational expense of Confluence is not significant due to the recursive reduction in the size of the set of bounding boxes, it may still be suboptimal for applications that prioritize speed. Thus, we also propose the non-IoU NMS algorithm Confluence-NMS (C-NMS).

**Algorithm 2** Confluence NMS
**Input:** $I(b,s) = \{b_0 : s_0, .., b_n : s_n\}, C_t$
 $I$ is a dictionary mapping a box $b_i$ to its corresponding confidence score $s_i$.
 $C_t$ is the Confluence threshold (optimal range: 0.6-0.9)
**Output:** $D$ is the final detection set
**begin:**
1: $D \leftarrow \{\}$
2: **while** $I \neq empty$ **do**
3: $b_m \leftarrow argmax(I)$
4: $D \leftarrow D \cup b_m, I \leftarrow I - b_m$
5: **for** $b_i$ in $I$ **do**
6: $normalise(b_m, b_i)$
7: $P \leftarrow P(b_m, b_i)$
8: **if** $P < C_t$ **then**
9: $I \leftarrow I - b_i$
10: **end if**
11: **end for**
12: **end while**
13: **return** $D$

Like other NMS variants, C-NMS retains the bounding box with the maxima score, however its performance in minimizing false positives whilst maximizing true positives is significantly improved. This is because it relies on Confluence to suppress false positives and retain true positives, improving both recall and precision. Algorithm 2 provides pseudo-code illustrating C-NMS.

C-NMS operates as follows:
1. In a dictionary $I$, mapping bounding boxes to their corresponding confidence scores, find the box with the highest score (line 3).
2. Retain the maxima $b_m$ reserving it in the final detections set $D$ and removing it from $I$ (line 4).
3. For each remaining box $b_i$, normalize its relationship with the maxima box $b_m$ (line 6) and

TABLE 1
COMPARATIVE PERFORMANCE OF CONFLUENCE AGAINST NMS ON MS-COCO

|  | Average Precision (area = all) (maxDets=100) IoU - variable | | | Average Precision (IoU: 0.50:0.95) (maxDets=100) Area - variable | | | Average Recall (IoU=0.50:0.95) (area = all) maxDets - variable | | | Average Recall (IoU=0.50:0.95) (maxDets=100) area - variable | | |
|---|---|---|---|---|---|---|---|---|---|---|---|---|
| Method | 0.50:0.95 | 0.50 | 0.75 | Small | Medium | Large | 1 | 10 | 100 | Small | Medium | Large |
| G-NMS | 0.414 | 0.573 | 0.446 | 0.225 | 0.355 | 0.643 | 0.361 | 0.459 | 0.459 | 0.231 | 0.402 | 0.700 |
| S-NMS-L | 0.423 | 0.584 | 0.464 | 0.230 | 0.385 | 0.644 | 0.361 | 0.496 | 0.497 | 0.255 | 0.464 | 0.725 |
| S-NMS-G | 0.435 | 0.598 | 0.477 | **0.235** | 0.391 | 0.663 | 0.361 | 0.504 | 0.507 | 0.261 | 0.468 | 0.741 |
| C-NMS | 0.437 | **0.600** | 0.481 | **0.235** | 0.390 | 0.662 | 0.361 | 0.507 | 0.512 | 0.264 | 0.470 | 0.746 |
| C-NMS-G | **0.439** | **0.600** | **0.482** | **0.235** | 0.393 | **0.665** | 0.361 | **0.515** | **0.520** | **0.268** | 0.478 | **0.757** |
| Confluence | 0.433 | 0.590 | 0.479 | 0.228 | **0.395** | 0.662 | **0.365** | 0.506 | 0.513 | 0.254 | **0.483** | 0.751 |

Performance of RetinaNet on the MS-COCO mini-val dataset with Greedy NMS (G-NMS), Soft-NMS linear (S-NMS-L), Soft-NMS gaussian (S-NMS-G), Confluence-NMS (C-NMS) and Confluence. The Greedy and Soft NMS variants were tested with an IoU threshold of 0.3, while an MD threshold of 0.7 was used for Confluence and Confluence NMS. The Confluence algorithms significantly outperform the IoU-based G-NMS and S-NMS in both Average Precision (AP) and Average Recall (AR) metrics.

TABLE 2
PER CLASS COMPARISON OF CONFLUENCE AGAINST NMS ON MS-COCO

| Method | Person | Car | Airplane | Train | Truck | Stop sign | Bird | Dog | Cat | Zebra | Backpack | Umbrella | Skis | Fork | Banana | Carrot | Cake | Potted plant | Laptop | Cell phone |
|---|---|---|---|---|---|---|---|---|---|---|---|---|---|---|---|---|---|---|---|---|
| G-NMS | 40.91 | 38.38 | 63.56 | 50.50 | 19.71 | 95.05 | 42.77 | 80.00 | 90.00 | 50.50 | 17.04 | 82.52 | 34.55 | 40.40 | 15.11 | 46.93 | 90.00 | 49.80 | 38.31 | 9.36 |
| S-NMS | 43.70 | 36.76 | 63.56 | 50.50 | 20.63 | 95.05 | 43.20 | 80.00 | 90.00 | 80.20 | 16.88 | 82.52 | 34.55 | 30.50 | 18.70 | 47.13 | 90.00 | 49.80 | 38.31 | 9.36 |
| C-NMS | 44.56 | 38.83 | 63.56 | 50.50 | 23.14 | 95.05 | 43.42 | 80.00 | 90.00 | 83.50 | 17.43 | 83.11 | 34.55 | 40.40 | 28.37 | 47.72 | 90.00 | 49.80 | 38.31 | 9.36 |
| Confluence | 44.21 | 36.96 | 63.56 | 50.50 | 21.45 | 90.10 | 43.09 | 80.00 | 90.00 | 80.20 | 17.85 | 82.52 | 30.40 | 42.97 | 24.70 | 47.68 | 90.00 | 49.80 | 37.67 | 9.36 |

AP results on 20 randomly selected COCO classes. C-NMS outperforms G-NMS and S-NMS strongly for classes that are often present in highly occluded or dense images, whereas performance on images characterised by lack of occlusion, low-density or lone objects is similar to NMS.

then calculate its proximity score $P(b_m, b_i)$ (line 7).

4. If $P(b_m, b_i)$ is less than or equal to the user defined Confluence threshold $C_t$ (line 8) remove box $b_i$ from the dictionary $I$ (line 9).
5. Repeat this process until all boxes have been retained in the set $D$ and/or suppressed.

Both Confluence and C-NMS have been implemented such that bounding box suppression can be achieved linearly or using Gaussian weighting. This is inspired by the use of Gaussian weighting by [7] to improve recall by down-sampling rather than suppressing high-scoring bounding boxes that overlap with optimal boxes.

## 4 DATASETS, ALGORITHMS AND EVALUATION

Experimental results presented in this paper were collected on the publicly available 2017 MS-COCO mini validation (mini-val) dataset [56] and 2018 CrowdHuman [57] validation datasets. These datasets were chosen to demonstrate Confluence on widely recognized standard and high-density, high-occlusion benchmarks.

The COCO mini-val set contains 5000 images of 80 classes. The CrowdHuman validation set contains 4370 images of people in highly crowded settings, with an average of 23 humans per image [57]. The AP calculations were obtained using the COCO-style evaluation metrics via the standard COCO API, using default settings including 100 maximum detections per image.

The G-NMS and S-NMS algorithms used were implemented by [21] and are publicly available on their GitHub repository.[2] These algorithms were evaluated using two state-of-the-art object detectors; RetinaNet-ResNet50 [5], and Mask-RCNN (HRNet) [3]. We used a Mask R-CNN implementation (and associated pretrained model) published by [81], which is publicly available on GitHub.[3]

The Mask-RCNN model used was trained by [81] on the CrowdHuman training set as described in [81]. The RetinaNet implementation and pretrained model used was obtained from GitHub.[4] It was trained on the MS COCO 2017 dataset as outlined on the repository. We did not conduct further model training, instead selecting publicly available pre-trained models in all cases. We simply replaced the default G-NMS module with the S-NMS implementation provided by [21], and our implementations of Confluence and C-NMS, gathering all results using default settings.

Threshold sensitivity analysis was also conducted to demonstrate the robustness of the Confluence threshold ($C_t$), by examining changes in AP attained by G-NMS and S-NMS over variations in IoU threshold in comparison to change in AP achieved by Confluence across variations in the $C_t$ threshold. Sensitivity analysis was achieved using RetinaNet, applied to the MS-COCO mini-val dataset.

---

[2] github.com/bharatsingh430/soft-nms
[3] github.com/hasanirtiza/Pedestron
[4] github.com/fizyr/keras-retinanet

TABLE 3
AP AND AR RESULTS – CROWDHUMAN AND MASK-RCNN

| Method | Average Precision (area = all) (maxDets=100) IoU - variable | | | Average Precision (IoU: 0.50:0.95) (maxDets=100) area - variable | | | Average Recall (IoU=0.50:0.95) (area = all) maxDets - variable | | | Average Recall (IoU=0.50:0.95) (maxDets=100) area - variable | | |
|---|---|---|---|---|---|---|---|---|---|---|---|---|
| | 0.50:0.95 | 0.50 | 0.75 | Small | Medium | Large | 1 | 10 | 100 | Small | Medium | Large |
| G-NMS | 0.470 | 0.741 | 0.505 | 0.435 | 0.545 | 0.443 | **0.033** | 0.257 | 0.519 | 0.524 | 0.603 | 0.485 |
| S-NMS-L | 0.498 | 0.787 | 0.533 | 0.443 | 0.571 | 0.476 | **0.033** | 0.260 | 0.567 | 0.548 | 0.646 | 0.534 |
| S-NMS-G | 0.479 | 0.778 | 0.506 | 0.435 | 0.559 | 0.452 | **0.033** | 0.254 | 0.570 | 0.557 | 0.659 | 0.531 |
| C-NMS | **0.508** | 0.799 | 0.549 | 0.447 | **0.578** | **0.485** | 0.033 | **0.261** | **0.591** | **0.569** | **0.669** | **0.557** |
| C-NMS-G | **0.508** | **0.800** | **0.550** | 0.447 | **0.578** | **0.485** | 0.033 | **0.261** | 0.590 | 0.567 | 0.668 | **0.557** |
| Confluence | 0.507 | 0.797 | 0.547 | **0.448** | 0.577 | **0.485** | **0.033** | **0.261** | 0.588 | 0.566 | 0.662 | 0.556 |

Performance of Mask-RCNN on the CrowdHuman validation dataset with Greedy NMS (G-NMS), Soft-NMS linear (S-NMS-L), Soft-NMS gaussian (S-NMS-G), Confluence-NMS (C-NMS), Confluence-NMS gaussian (C-NMS-G) and Confluence. The Greedy and Soft NMS variants were tested with an IoU threshold of 0.5, while an $C_t$ threshold of 0.5 was used for the Confluence algorithms.

## 5 RESULTS

In this section, we provide performance results on both MS-COCO mini-val and CrowdHuman datasets. In Table 1 we compare the performance of Confluence and C-NMS against G-NMS and S-NMS on the MS-COCO mini-val dataset using RetinaNet. The Confluence $C_t$ threshold was set to 0.7. It is clear that both Confluence and C-NMS improve object detector performance, particularly on the more stringent AP@0.5:0.95 AP calculation. When compared to G-NMS, S-NMS gaussian (S-NMS-G) and S-NMS linear (S-NMS-L), Confluence achieves gains in AP of 1.0-1.9% while C-NMS outperformed G-NMS and S-NMS-L by 1.4-2.3%. Similarly, at the PASCAL VOC AP metric of IoU@0.5, Confluence outperforms G-NMS and S-NMS-L by 0.6-1.7% while C-NMS outperforms G-NMS and S-NMS-L by 1.6-2.7%.

Significant improvements in AR are also achieved by Confluence and C-NMS, with gains of 1.5-5.3% at max detections per image of 100. Confluence and C-NMS also outperform G-NMS and S-NMS-L across all object sizes by up to 3.5% AP and 8.1% AR.

Gaussian confidence score decaying is more computationally expensive, and only results in gains in AP and AR when a very low confidence threshold is used, reducing its practical applicability. Regardless, we provide results for C-NMS-G, which outperforms S-NMS-G by 0.4% on the AP@0.5-0.95 metric and 0.2% on the PASCAL VOC 0.5% AP metric. Gains in AR range from 1.3% on the max detections of 100 metric, and gains of up to 1.6% on objects of varying sizes. These improvements are significant for the MS-COCO dataset and evaluation metric.

Table 2 presents RetinaNet performance across 20 randomly selected classes in the MS-COCO dataset. Confluence and C-NMS outperform G-NMS and S-NMS strongly for classes that are often present in highly occluded or dense images, such as person, truck, zebra, and banana, where improvements ranged from 2.51-9.67%. Gains in AP for object detection on classes that are often represented as lone objects and not occluded, for example laptop, dog, cat, and airplane, are insignificant and often equivalent to AP achieved by G-NMS and S-NMS.

Table 3 presents the same experiments performed using Mask-RCNN evaluated on the CrowdHuman dataset. On the AP@0.5:0.95 metric, Confluence achieved gains in AP of 1.2-4%, while C-NMS outperformed G-NMS and S-NMS-L by 1.0-3.8%. Similarly, at the PASCAL VOC AP metric of IoU@0.5, Confluence improves AP by 1.4-6% while C-NMS outperforms G-NMS and S-NMS-L by 1.2-5.8%. Significant gains in AR are also achieved with gains of 2.4-7.3% at max detections per image of 100. Confluence and C-NMS also outperforms G-NMS and S-NMS-L across all object sizes by up to 4.4% AP and 7.4% AR. C-NMS-G significantly outperforms S-NMS-G by 2.9% on the AP@0.5-0.95 metric and 2.2% on the PASCAL VOC 0.5% AP metric. AR improves by 2.0% on the max detections of 100 metric, and up to 2.6% on objects of varying sizes.

Notably, improvements in AP and AR were greater on CrowdHuman and MS-COCO classes that frequently involve high occlusion. This suggests that the task of bounding box retention and removal can better be achieved by interpreting confluent bounding boxes according to coherence of borders rather than overlap.

### 5.1 Sensitivity Analysis

This subsection presents the results of threshold sensitivity analysis comparing the sensitivity of the G-NMS and S-NMS IoU threshold to the Confluence and C-NMS $C_t$ threshold. High sensitivity is detrimental due to the consequently high fluctuations in AP and AR when the algorithm is confronted with variations in image density and occlusion within a dataset. The results shown in Fig. 8 were collected using RetinaNet and the MS-COCO mini-val dataset, evaluated using the AP@0.5:0.95 metric.

Both parameters were varied incrementally by 0.1. The optimal threshold for G-NMS and S-NMS on MS-COCO lies between 0.3-0.6. This is because a high IoU threshold results in only highly overlapping bounding boxes being removed, resulting in a high number of false positives but greater recall. In contrast, a lower IoU threshold removes more bounding boxes, reducing recall yet minimising false positives.

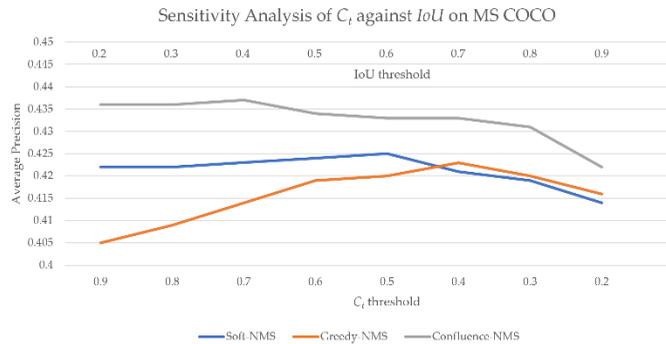

Fig. 8. Sensitivity of RetinaNet to the Confluence and IoU hyper-parameters.

The optimal range for Confluence and C-NMS ranges between 0.5 and 0.8. This is because a high $C_t$ value indicates low proximity of bounding box borders, while lower values indicate greater bounding box confluence. Thus, the higher the $C_t$ threshold, the more bounding boxes are removed. This is why the optimum threshold value for Confluence and C-NMS is higher than that used by the IoU dependent G-NMS and S-NMS.

Performance by both algorithms tends to decrease outside these ranges. Note that the variation in AP for Confluence and C-NMS within its optimal range is more stable than G-NMS and S-NMS, with variation in AP for $C_t$ being 0.3% while variation in AP for IoU is 0.6%. This means that the $C_t$ threshold is less sensitive to fluctuations in object density and occlusion, which makes it more robust. Furthermore, as shown by Fig. 8 performance of Confluence and C-NMS always remains approximately 1.5% better than G-NMS and S-NMS, even at the optimal IoU threshold.

## 6 DISCUSSION

This section will relate the quantitative results presented in Section 5 to a qualitative comparison of Confluence, C-NMS and the IoU-based G-NMS and S-NMS. It will explain how and why Confluence returns optimal bounding boxes, using qualitative data to demonstrate why coherence of bounding box borders is a more appropriate metric than IoU to use in bounding box selection and suppression in object detection. We will also provide insight into possible future work to further improve the performance and applicability of the Confluence algorithm.

### 6.1 Qualitative Comparison of Confluence with IoU-based G-NMS and S-NMS

A fundamental problem with IoU based suppression of bounding boxes is the elimination of true positives in high density images. Once the highest confidence bounding box $b$ is selected, any detection with a sufficiently high overlap with $b$ is removed. In situations where objects are occluded by other objects of the same class, for example, when a person occludes another person as shown in Fig. 1 and 9, high IoU will often result in the suppression of detections denoting true positives.

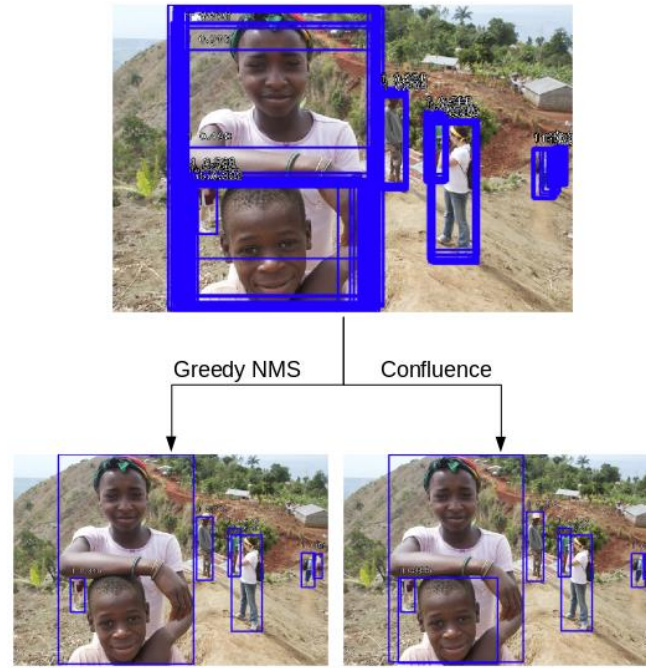

Fig. 9. G-NMS labelled the two people in the foreground as one person whilst Confluence distinguishes them as two separate objects due to the two areas of bounding box confluence.

The raw, unfiltered object detector output is illustrated in Fig. 1 and 9 at a confidence threshold of 5%. It is evident by the thick confluence of proposals that all objects are detected and localized correctly. Thus, the aim of the post-processing stage is to maximize precision by selecting an optimal detection to represent each true positive, without lowering recall by suppressing true positives. IoU variants of NMS, such as G-NMS and S-NMS do not achieve this when applied to these images. Their reliance on the maxima confidence score causes them to return suboptimal bounding boxes, while their IoU dependency causes them to suppress true positives. In contrast, Confluence uses the heavy cluster of bounding boxes as an indicator of the presence of an object, thus returning one bounding box per cluster. This results in both higher recall and precision.

Unlike G-NMS and S-NMS, Confluence rewards high confluence, interpreting clustering as a vote of confidence by the neural network on the likeliest location of an object in an image. Thus, perhaps the most effective means by which Confluence can be evaluated is by qualitative data. Fig. 9, 10, 11, 12, 13 and 14 illustrate qualitative results using the output of RetinaNet on images from the MS-COCO dataset.

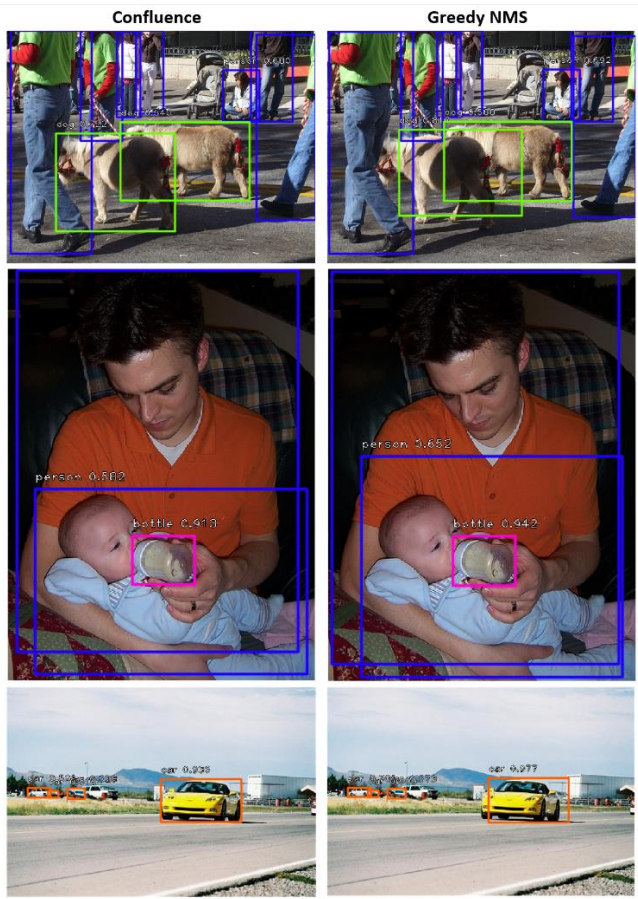

Fig. 10. Even when objects are not occluded, or highly proximate, the highest scoring box is not always optimal. This is true for all classes.

*1) Improved Bounding Box Selection:* As previously discussed, object detectors return many bounding boxes, and their corresponding scores in locations where the probability of an object being present is high. Although most NMS variants interpret the bounding box with the highest confidence score within a cluster as the optimal bounding box, there are many instances where the highest scoring box is not the optimal bounding box. A few examples of these situations are provided in Fig. 10 and 11.

Many of these instances involved people, at close proximity, such as those shown in Fig. 11. G-NMS and S-NMS simply return the highest scoring box, even if it is too large or too small. In contrast, Confluence has the capacity to return a more accurate box by taking advantage of the confluence of tighter fitting bounding boxes around each person.

Rather than simply selecting the highest scoring bounding box, Confluence selects the bounding box which is the most coherent with every other bounding box in the cluster. Consequently, if a highly confluent bounding box is a better representation of the object, it will be returned by Confluence despite having a lower confidence score. This improvement in bounding box selection is most evident when the object detector is a proposal based DCNN, such as RetinaNet, which has a tendency of returning very dense confluence around objects.

*2) Improved Bounding Box Selection Improves Recall:* The reliance of NMS on the classification confidence score as the sole means by which an optimal bounding box is selected reduces recall. For example, Fig. 12 illustrates how G-NMS selects the highest confidence box (87.7% confidence, shown in red) to locate the boy, but this bounding box is sub-optimal in comparison to the bounding box selected by Confluence (82.1% confidence). Due to G-NMS's selection of the maxima, it suppressed the bounding box allocated to the man standing behind the boy (shown in yellow), as they share a high IoU, thus reducing recall. In contrast, Confluence retains both bounding boxes locating both the man and boy.

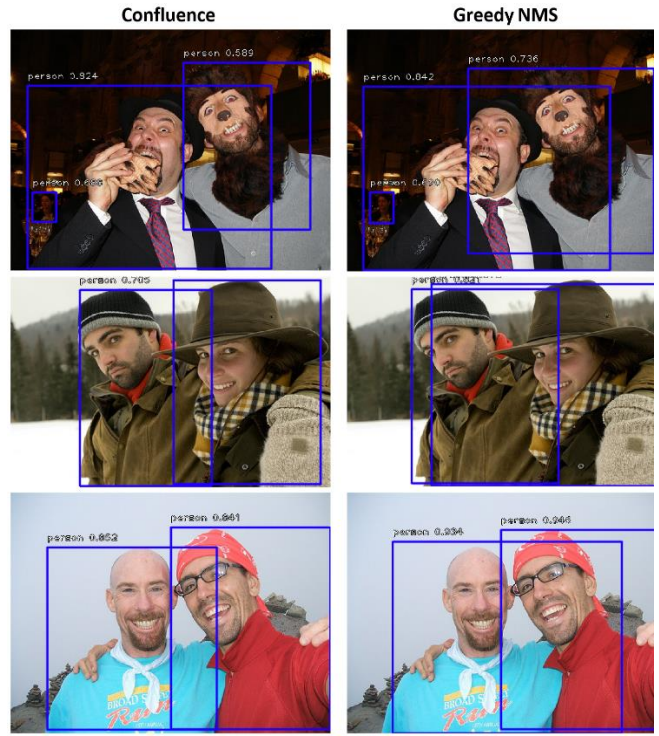

Fig. 11. The optimal box does not always have the highest confidence score. Often, a bounding box which is too large is returned in instances where objects, such as people are at close proximity.

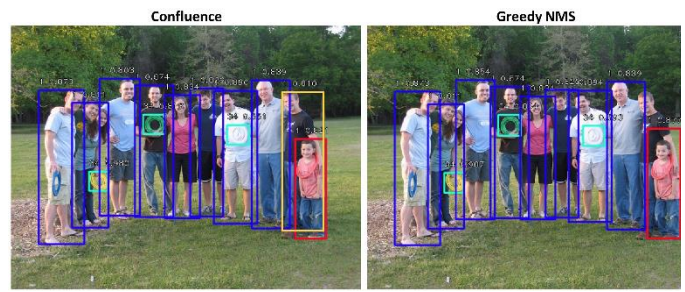

Fig. 12. Reduction in Recall. Note the poor selection by G-NMS of a bounding box to locate the boy on the far right (red box), and the suppression of the bounding box allocated to the man standing behind him (yellow box).

*3) Suppression of False Positives via Manhattan Distance Improves Accuracy:* It was observed that due to suppression of false positives via IoU, in some situations, G-NMS suppresses a bounding box which has a high IoU with a higher confidence box, even if it correctly locates a second object. NMS is then forced to select a sub-optimal bounding box to locate the second object, due to suppression of the optimal box. For example, Fig. 13 shows the output of G-NMS, and Confluence on the same image. G-NMS suppresses the optimal, high confidence (66%) box allocated to the giraffe in the background due to its high IoU with the high confidence (94.8%) box allocated to the giraffe in the foreground. It is then forced to return a low confidence (50.7%), sub-optimal bounding box for the giraffe in the background.

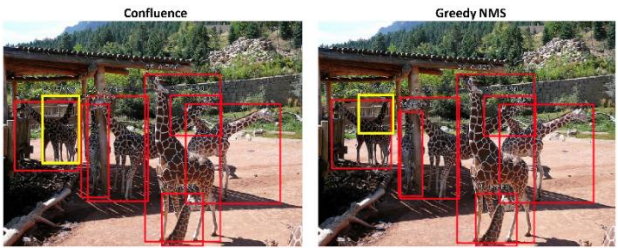

Fig. 13. The optimal bounding box for the background giraffe on the far left (yellow) is suppressed due to IoU, which forces G-NMS to return a sub-optimal bounding box.

This issue can be rectified by increasing the NMS IoU overlap threshold, however this comes at a cost – the number of false positives returned increases significantly. In contrast, the Confluence algorithm uses areas of confluence returned by the object detector to determine whether or not a second object is present, thus making it more robust to this issue. It has the capacity to return optimal bounding boxes without returning false positives.

*4) Improved Recall:* Confluence achieves better recall than G-NMS because bounding box removal is based on the coherence of bounding box borders with each other, rather than the extent of their overlap. The benefits of this approach are most evident when objects are occluding each other, for example, when a person is standing in front of another person, as shown in Fig. 1, 8 and 12. Although the smaller bounding box is a subset of the larger box, its borders are not sufficiently coherent to be removed by Confluence. However, they share a large overlap, which means the lower confidence box is removed. This suggests that Confluence is more robust to high occlusion and explains why its recall is higher than G-NMS on all tested object detectors.

One shortcoming of the use of Confluence to remove false positives is its tendency to retain those false positives which are not confluent with other boxes in a given cluster. This can be seen in Fig. 14. The tennis racket is surrounded by two bounding boxes rather than one (see annotation 1), as the two boxes are not confluent. In situations like this, the object detector is not confident, which causes it to return spurious bounding boxes around an object. In these (uncommon) cases, NMS has higher precision, as it harshly removes any highly overlapping bounding boxes.

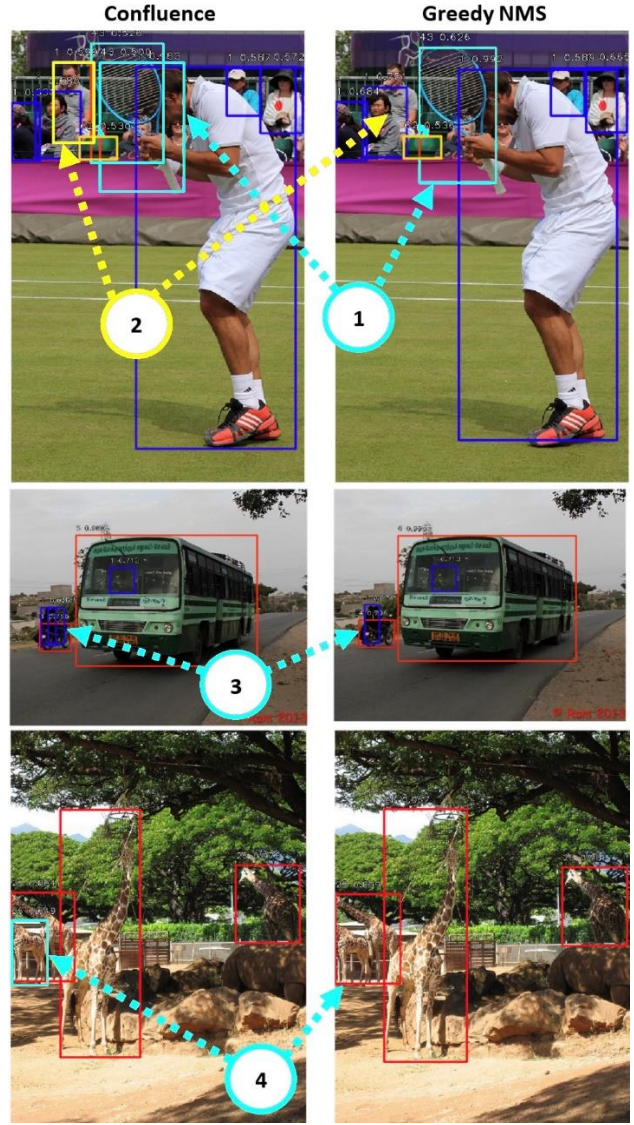

Fig. 14. The giraffe numbered 4 is a subset of the bounding box allocated to the giraffe in the background. It is removed by NMS, yet retained by Confluence, accounting for its higher recall. This also occurs in crowds, as shown by 2, and where objects are occluding each other, as shown by 3.

## 6.2 Future Work

Our experimental results strongly indicate that Confluence and Confluence NMS are superior alternatives to IoU-based variants of NMS including the de-facto standard Greedy NMS and the state-of-the-art Soft NMS in both multi-class and single-class object detection. However, the extent of its improvement over G-NMS and S-NMS may have been obscured by the way that the MS-COCO AP calculation processes bounding boxes. In AP calculators, bounding boxes are ranked by confidence score [43], [44], [9]. Consequently, if a sub-optimal box has a higher confidence score than a superior lower confidence box, the sub-optimal bounding box will be chosen by the AP calculator, and the superior box chosen by Confluence will be ignored. This degrades the overall AP score. Thus, future work could encompass the development of a confluence score to be used instead of the confidence score to rank bounding boxes in AP calculations. This would overcome the disadvantage faced by Confluence, enabling more effective evaluation of the extent of improvements in both bounding box retention and removal.

Furthermore, the hyper-parameter used by Confluence could be learned to optimize performance in any object detector. Another interesting area of research could be investigation of the applicability of the Manhattan Distance principles used by Confluence as a possible non-IoU alternative to Jaccard Index based regression modules, and AP calculators.

## 7 CONCLUSION

We have proposed Confluence, a non-IoU alternative to Non-Maxima Suppression and its variants. Confluence does not rely on IoU or the maxima classification confidence score in bounding box retention and suppression. Empirical results obtained using multiple object detectors on the MS COCO and CrowdHuman benchmarks attest to its superior performance when compared to the de-facto standard and state-of-the-art NMS variants.

Furthermore, threshold sensitivity analysis indicates that Confluence is more robust to occlusion. Finally, it can be seamlessly integrated within currently used object detectors without modifications or training, making it an attractive alternative to NMS variants in object detection. Confluence represents a paradigm shift away from the heavily researched and used IoU towards a more robust boundary distance metric, which may be used to replace IoU in other applications, ranging from regression modules to localization losses and AP calculators.

## ACKNOWLEDGMENT

This research was funded by an Australian Research Training Program (RTP) Scholarship.
Corresponding author: Andrew Shepley

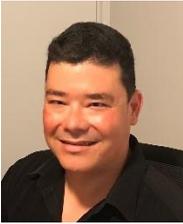

**Andrew Shepley** is a Lecturer and researcher at the University of New England (UNE), NSW, Australia. He received a Ph.D. in Computer Science at UNE. He received a BSc. Hons at the University of New South Wales (UNSW), a Grad. Degree Information Science at UNSW, and a Grad Dip Education at UNE. His research experience is centered on improving performance of computer vision by developing algorithms to improve the accuracy and efficiency of deep learning systems and artificial intelligence. His current research involves the development of biometrics for individual recognition.

**Gregory Falzon** is Associate Professor of Precision Agriculture Systems at Flinders University, Adelaide, Australia. He also holds Adjunct Associate Professor status and was previously a Senior Lecturer in Computational Science at the University of New England in Armidale, Australia. He holds a PhD in biomedical image analysis , and a Grad Dip Education and BSC (Hons I) from the University of New England. Major professional achievements include the 2020 ATSE ICM Agrifood Award for advancing AI in agriculture; President's Medal, ASCCT; Science & Innovation Award, Australian Government Department of Agriculture, and a solid track-record of industry funding. Current research spans the development of novel technologies especially machine vision systems and their applications to agriculture. He has been a member of IEEE since 2011.

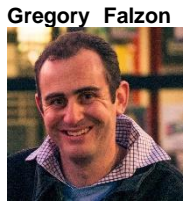

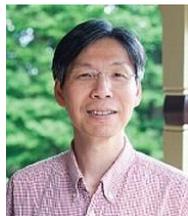

**Paul Kwan** is a Professor of Computer Science at The University of New England in Armidale, Australia. He has a PhD in Advanced Engineering Systems, specialized in Intelligent Interaction Technologies, from The University of Tsukuba in Japan, a BSc and an MSc degree, both in Computer Science, from Cornell University and University of Arizona in the United States. His research fields include Artificial Intelligence, Computer Vision, Computational Modelling, Image Processing and Machine Learning, and their applications to digital agriculture, human and animal biometrics, computer simulation and analysis of animal diseases and invasive pests spread. He has been a Member of Australian Computer Society since 2006, a Senior Member of Association of Computing Machinery since 2008, and a Senior Member of IEEE since 2010.

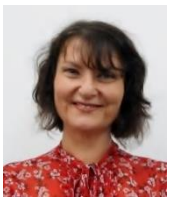

**Ljiljana Brankovic** is a Professor in Computational Science at the University of New England (UNE), Australia. She received her Graduate Electrical Engineer degree from the University of Belgrade, Serbia, and her PhD in Computer Science from the University of Newcastle, NSW, Australia. Her main research areas include Algorithms, Cybersecurity and Privacy, and Graph Theory and Combinatorics. She is a Fellow of the Institute of Combinatorics and its Applications and a Life Member of the Combinatorial Mathematics Society of Australasia. From 2007 to 2020, she served as an Australian Computer Society Chair of the National Committee for Computer Security.